\documentclass[letterpaper, 10 pt, journal, twoside]{ieeetran}
\IEEEoverridecommandlockouts

\usepackage[style=ieee,sorting=none,backend=bibtex]{biblatex}
\usepackage[dvipsnames]{xcolor}

\usepackage{lipsum}
\usepackage{wrapfig}
\usepackage{subfig}
\usepackage[pdftex]{graphicx}
\usepackage{amsmath}
\usepackage{amsfonts}
\usepackage{multirow}
\graphicspath{{./images/}}

\title{Recent Advances of Deep Robotic Affordance Learning: A Reinforcement Learning Perspective}

\author{Xintong Yang$^{1}$, Ze Ji$^{1}$, Jing Wu$^{2}$, and Yu-kun Lai$^{2}$%
\thanks{Manuscript under-review \textit{(Corresponding author: Ze Ji)}.}
\thanks{$^{1}$School of Engineering, Cardiff University, Cardiff, UK{\texttt{\small} \{yangx66, jiz1\}@cardiff.ac.uk}}
\thanks{$^{2}$School of Computer Science and Informatics, Cardiff University, Cardiff, UK{\texttt{\small} \{wuj11, laiy4\}@cardiff.ac.uk}}%
}

\begin{document}

\maketitle
\thispagestyle{empty}
\pagestyle{empty}

\begin{abstract}
As a popular concept proposed in the field of psychology, affordance has been regarded as one of the important abilities that enable humans to understand and interact with the environment. Briefly, it captures the possibilities and effects of the actions of an agent applied to a specific object or, more generally, a part of the environment. This paper provides a short review of the recent developments of deep robotic affordance learning (DRAL), which aims to develop data-driven methods that use the concept of affordance to aid in robotic tasks. We first classify these papers from a reinforcement learning (RL) perspective, and draw connections between RL and affordances. The technical details of each category are discussed and their limitations identified. We further summarise them and identify future challenges from the aspects of observations, actions, affordance representation, data-collection and real-world deployment. A final remark is given at the end to propose a promising future direction of the RL-based affordance definition to include the predictions of arbitrary action consequences.
\end{abstract}

\section{Introduction}
Humans interact with various objects in the environment in a purposeful and meaningful way, because we have the ability to understand affordances -- the functionalities of objects, the possibilities and effects of our actions and the relationship between the two. As originally defined by Gibson \cite{gibson1982concept}, the affordances of an object or a place in an environment provide knowledge about what actions are possible and what the consequences of these actions are with respect to a certain agent (a human, an animal or a robot). In short, it indicates \textbf{possibilities} and \textbf{effects} of the agent's actions given an object or a part (an image observation) of the environment. In the field of robotics, affordances could serve with great potential to bridge robot perception and action \cite{good2007affordances}. This has been actively integrated and explored with machine learning techniques in recent years \cite{jamone2016affordances, yamanobe2017brief, ardon2021building, hassan2021affordance}. \citeauthor*{jamone2016affordances} proposed a thorough review and drew connections among the studies of affordances in psychology, neuroscience and robotics \cite{jamone2016affordances}. \citeauthor*{yamanobe2017brief} summarised the use of affordances specifically in robotic manipulation tasks \cite{yamanobe2017brief}. \citeauthor*{ardon2021building} summarised and provided guidance on design choices and how affordance relations can be used to boost policy learning \cite{ardon2021building}.

However, as pointed out in \cite{hassan2021affordance}, there is still a lack of consensus for a formal definition of affordances, and many previous works are limited in the cases with object-affordances. Inspired by a recent works that proposed to define, learn and compute affordances in reinforcement learning (RL) on Markov decision processes (MDPs) of any kind \cite{khetarpal2020can}, we propose in this paper to summarise and classify recent publications (since 2015) in deep robotic affordance learning (DRAL) following the RL-based definition. There are several motivations to do so:

\begin{itemize}
    \item The RL-based definition helps to unify and classify DRAL works from a behavioural learning perspective, providing new insights to understand and clarify the different usages of affordances in the literature;
    \item The definition in \cite{khetarpal2020can} is the most general in the literature as all concepts are defined over a generic MDP without any assumption of the environmental or agent aspect. It suits any kind of environmental affordances and agents as long as they can be described by MDPs, which is commonly achievable.
    \item As the primary aim of DRAL is to enable robots to infer afforded actions, the RL community provides a rich body of methods ready to be integrated with affordances;
    \item Understanding and analysing the concept based on a mathematical framework helps to provide computationally and practically valuable insights;
\end{itemize}

In practice, knowing the affordances means knowing the desired effects of some actions and whether these effects can be realised at some situations. With this in mind, \citeauthor*{khetarpal2020can} introduced the notion of \textit{intents} that captures the desired outcome of an action based on the reinforcement learning (RL) framework \cite{sutton2018reinforcement, khetarpal2020can}. For example, the intent of a moving right action in a gridworld task is the agent being moved to the cell on the right. The intent is not always satisfied, e.g., when the cell on the right is a wall. Thus, the definition of affordances is a subset of the state-action space in which the intent is indeed satisfied \cite{khetarpal2020can}. In order words, the moving right action is afforded at every state where the moving right intent is satisfied. 

Notice that there are two levels of the topic: 1) the learning and discovery of affordances and 2) the use of affordances. Researchers have only started recently to study the first level, e.g., option/subgoal discovery \cite{wulfmeier2021data}. Most research focuses on the use of the knowledge of affordances, meaning how to estimate the action possibilities and/or infer the afforded actions. These works are classified into three categories as follow.

\begin{itemize}
    \item For the majority of the DRAL works, the focus is to estimate the action possibilities given an observation and then infer afforded actions from it (Section~\ref{sec:MAP}). These works can be further classified into methods that model the action possibilities as binary variables (subsection~\ref{subsec:ALIS}) \cite{myers2015affordance, kokic2017affordance, nguyen2017object, do2018affordancenet, chu2019toward, mandikal2020graff, hamalainen2019affordance, borja2022affordance} and continuous variables (subsection~\ref{subsec:AS}) \cite{zeng2018robotic, cai2019metagrasp, wu2020learning, yang2021learning, mo2021where2act, ichter2022do};
    \item The second line of works propose to generate afforded actions from a set of object keypoints (Section~\ref{sec:KA}) \cite{manuelli2019kpam, gao2021kpam2, gao2021kpamsc, qin2020keto, turpin2021gift, xu2021affordance}. The keypoints were used to geometrically constrain the search space of action inference methods within the set of afforded actions.
    \item The last part of the reviewed papers suggest to learn a partial dynamic model for only afforded actions, resulting in faster model learning and motion planning (Section~\ref{sec:MAPE}) \cite{khetarpal2020can, xu2020daf, khetarpal2021temporally}. 
\end{itemize}

The rest of this review is organised as follow. Section~\ref{sec:ADMDP} briefly recalls the definition of affordances in reinforcement learning proposed by \cite{khetarpal2020can}, classifies the reviewed works and draws connections between RL and affordances. Section~\ref{sec:MAP}, \ref{sec:MAPE} and \ref{sec:KA} provides the main technical ideas and discusses the pros and cons of the reviewed papers. Section~\ref{sec:discussion} summarises these works and poses future challenges from the perspectives of observations, actions, affordance representations, data collection and rel-world deployment. Section~\ref{sec:Concl} concludes this review.

\section{Affordance definition in MDPs}\label{sec:ADMDP}

For the sake of clarity, we recall in this section the reinforcement learning (RL) problem and the definition of affordance based on the Morkov Decision Processes (MDPs) \cite{khetarpal2020can}. 

An MDP is a tuple $M=\langle \mathcal{S}, \mathcal{A}, r, P, \gamma \rangle$, where $\mathcal{S}$ is the set of states, $\mathcal{A}$ is the set of actions, $r$ is the reward function, $P(s'|s,a)$ is the system transition dynamics and $\gamma\in[0,1]$ is the discount factor \cite{sutton2018reinforcement}. The RL problem is in general to find an optimal policy, $\pi:\mathcal{S}\rightarrow\mathcal{A}$, which produces actions that maximise the expected discounted future return $\mathbb{E}_{\pi}[G_t]=\mathbb{E}_{\pi}\left[\sum^{\infty}_t\gamma^t r_t\right]$. The typical process of leaning such a policy loops over the procedures of data collection, policy evaluation and policy improvement \cite{sutton2018reinforcement}.

Given an action $a\in\mathcal{A}$, an intent $I_a(s)$ maps a state $s\in\mathcal{S}$ to a state distribution that the action is intended to achieve. The intent model can thus be seen as a partial dynamic model: $P_I(s'|s,a)$, which only captures the dynamics for a subset of states where the action has a desired effect. Given the full system dynamic model $P(s'|s,a)$, an intent is satisfied (i.e., an action is affordable) at a state, to a degree $\epsilon$, if and only if:

\begin{equation}\label{eq:affordable}
    d(P_I(s'|s,a), P(s'|s,a)) \leq \epsilon
\end{equation}

\noindent where $d$ is a function that measures the difference between two distributions and $\epsilon\in[0,1]$ is a precision parameter. Given a set of intents $\mathcal{I}=\cup_{a\in\mathcal{A}}I_a$, the affordance is then defined as a relation $\mathcal{AF_{I}}\subseteq \mathcal{S}\times\mathcal{A}$, such that $\forall (s,a)\in \mathcal{AF_{I}}$, Eq.~\ref{eq:affordable} is satisfied. Accordingly, an affordance prediction model, or an action possibility model, gives the probability of whether a pair of state and action belongs to the set of affordance:

$$
p^{\mathcal{AF}}(s,a)=p((s,a)\in \mathcal{AF_{I}})
$$

\textbf{Remark 1:} Practically speaking, knowing the affordance set means knowing the desired effects of a subset of actions (intents, action effects) and the subset of states that these effects can be achieved (states where the intents are satisfied, action possibilities). Before inferring the afforded actions or computing the action possibilities, one must know what actions, or what effects, are concerned or to be used. This logic implies that a robot must have learnt or been given some prior knowledge of the concerned actions beforehand. At the current stage of DRAL research, this knowledge was given by researchers, who then focused on the estimation of action possibilities and the inference of afforded actions. We categorise and discuss these methods in three classes:  

\begin{itemize}
    \item works that tried to infer the afforded actions from the estimated action possibilities $\hat{p}^{\mathcal{AF}}$ (section~\ref{sec:MAP});
    \item works that tried to infer the afforded actions of objects in terms of keypoints (section~\ref{sec:KA});
    \item works that tried to infer afforded actions by planning with $\hat{p}^{\mathcal{AF}}$ and a learnt partial dynamic model associated with intents, $\hat{P}_I(s'|s,a)$ (section~\ref{sec:MAPE}).
\end{itemize}

In the following sections, especially section~\ref{sec:MAP} and \ref{sec:KA}, the readers shall see that most recent works in using affordances in robotics did not reside their methods in the RL framework, although these methods can be explained from the RL perspective.

\textbf{Remark 2:} From the RL perspective, or a behavioural learning perspective, the knowledge of affordances can help to accelerate and improve almost every aspect of the RL process by constraining the action space. These include the learning of a value function, a policy, or a world model, the exploration direction, and the action inference process. For example, if an action possibility model is available, one can integrate it into the exploration process of any RL algorithm such that it only collects experiences where actions do cause changes to the environment. Alternatively, on may constrain the updates of a policy within the set of affordable actions. Also, as demonstrated by \cite{khetarpal2020can}, focusing on the set of afforded actions simplifies the learning of a world model and accelerate planning. 

Either for data collection, policy learning, world-model learning or action planning, the use of affordances in RL may have its best potential in the hierarchical reinforcement learning (HRL) framework where an agent learns to use a set of motion primitives (sub-policies, skills, temporal-extended actions) to achieve different tasks \cite{sutton1999between}. Knowing the possibilities and effects of the skills can accelerate learning by constraining and guiding the choices of exploring skills, filtering out experiences with irrelevant or non-effective actions, etc., reducing the lengthy exploration and learning processes for tasks with long horizons.

\textbf{Remark 3:} A further step to take in this regard is the learning and discovery of affordances. Knowing the set of affordances is promising and valuable in terms of accelerating learning, however, enabling an agent to learn and discover affordances makes the agent robust to potential changes in the environment and the agent itself. This is closely related to the popular topic of option/subgoal discovery in HRL \cite{wulfmeier2021data}. Future research topics in this regard include learning new skills, adapting old skills, skill composition, action space design, etc. One can envision a robot acquiring new skills in a new environment or modifying old skills as its hardware wear and tear.

\section{Modelling Action Possibilities}\label{sec:MAP}

This section discusses recent papers on modelling and learning action possibilities. This section examines two lines of works that represents $\hat{p}^{\mathcal{AF}}$ (whether an action or a set of actions is affordable given an observation) as binary segmentation masks (\ref{subsec:ALIS}) and continuous action success scores (\ref{subsec:AS}). We summarise these works and discuss their limitations in subsection~\ref{subsec:MAP-Limit}. 

Based on the definition given in section~\ref{sec:ADMDP}, these methods compute $\hat{p}^{\mathcal{AF}}$ for a set of actions given a state. The estimated $\hat{p}^{\mathcal{AF}}$ can be used to infer desirable actions in various ways based on its representations, such as taking the action with the maximum possibility, i.e., computing $argmax_{a\in\mathcal{A}}\  \hat{p}^{\mathcal{AF}}$. In practice, computing $\hat{p}^{\mathcal{AF}}$ is commonly based on sensory observations such as point clouds or images, instead of the true system states. The observation representations, training methods, deployment tasks and motion generation methods adopted by these works are summarised in TABLE~\ref{tab:action-prob-summary}.

\begin{table*}[t]
\caption{Summary of papers focused on learning action possibilities. \textbf{Cat.:} category; \textbf{IS:} image segmentation; \textbf{AS:} action scores; \textbf{PCD:} point cloud data; \textbf{SL:} supervised learning; \textbf{SSL:} self-supervised learning; \textbf{Sim:} simulation \textbf{Real:} real-world; \textbf{DoF:} degree of freedom; \textbf{PJG:} parallel-jaw grasp; \textbf{BOSM:} binary object segmentation mask; \textbf{RL:} reinforcement learning; \textbf{Sim-to-Real:} simulation to real world transfer.}    
\label{tab:action-prob-summary}
\begin{center}
\begin{tabular}{c|c|c|c|c|c|c}
\hline
Paper                        & Cat. & Affordance (afforded actions) & Input & Method & Deployment Task & Motion\\
\hline
\cite{myers2015affordance}      & IS & Created UMD dataset & RGBD        & SL & None & -\\
\cite{kokic2017affordance}      & IS & Grasp; Cut; Poke; Pound; Pour; Support & PCD & SL & 3-Finger Dexterous grasp (Sim) & Planning\\ 
\cite{nguyen2017object}         & IS & Created IIT-AFF dataset & RGB & SL & Dexterous grasp (Real) & Planning\\ 
\cite{do2018affordancenet}      & IS & from IIT-AFF\cite{nguyen2017object}; UMD\cite{myers2015affordance} datasets & RGB & SL & Dexterous grasp (Real) & Planning\\
\cite{chu2019toward}            & IS & from UMD\cite{myers2015affordance} dataset & RGBD & SL & 4DoF PJG; bean-scoop (Real) & Planning\\
\cite{mandikal2020graff}        & IS & Dexterous grasp & RGBD & SL & Dexterous grasp (Sim) & RL\\
\cite{hamalainen2019affordance} & IS & from UMD\cite{myers2015affordance} dataset & RGB(D) & SL & Cup-locate (Sim \& Real) & Planning\\
\cite{borja2022affordance}      & IS & Grasp & RGBD & SSL & 4DoF PJG (Sim \& Real) & Primitive \& RL\\
\hline
\cite{zeng2018robotic}          & AS & Grasp; Suction & RGBD & SL & 3DoF PJG \& suction (Real) & Primitive\\
\cite{cai2019metagrasp}         & AS & Grasp & RGB & SSL & 4DoF PJG (Sim \& Real) & Primitive\\
\cite{wu2020learning}           & AS & Grasp; Push & RGB & SSL & 4DoF PJG \& push (Sim \& Real) & Primitive\\
\cite{yang2021learning}         & AS & Grasp & BOSM & SSL & 4DoF PJG (Sim-to-Real) & Primitive \& RL\\
\cite{mo2021where2act}          & AS & Push; Pull & RGBD \& PCD & SSL & Push \& pull (Sim) & Primitive\\
\cite{ichter2022do}             & AS & Pick; Move; Place; Go-to; Open/close drawer & RGB & SL/RL & Kitchen tasks (Sim \& Real) & Primitive \& SL/RL\\
\hline
\end{tabular}
\end{center}
\end{table*}

\subsection{Image segmentation}\label{subsec:ALIS}

\begin{figure}[t]
    \centering
	\includegraphics[width=0.48\textwidth]{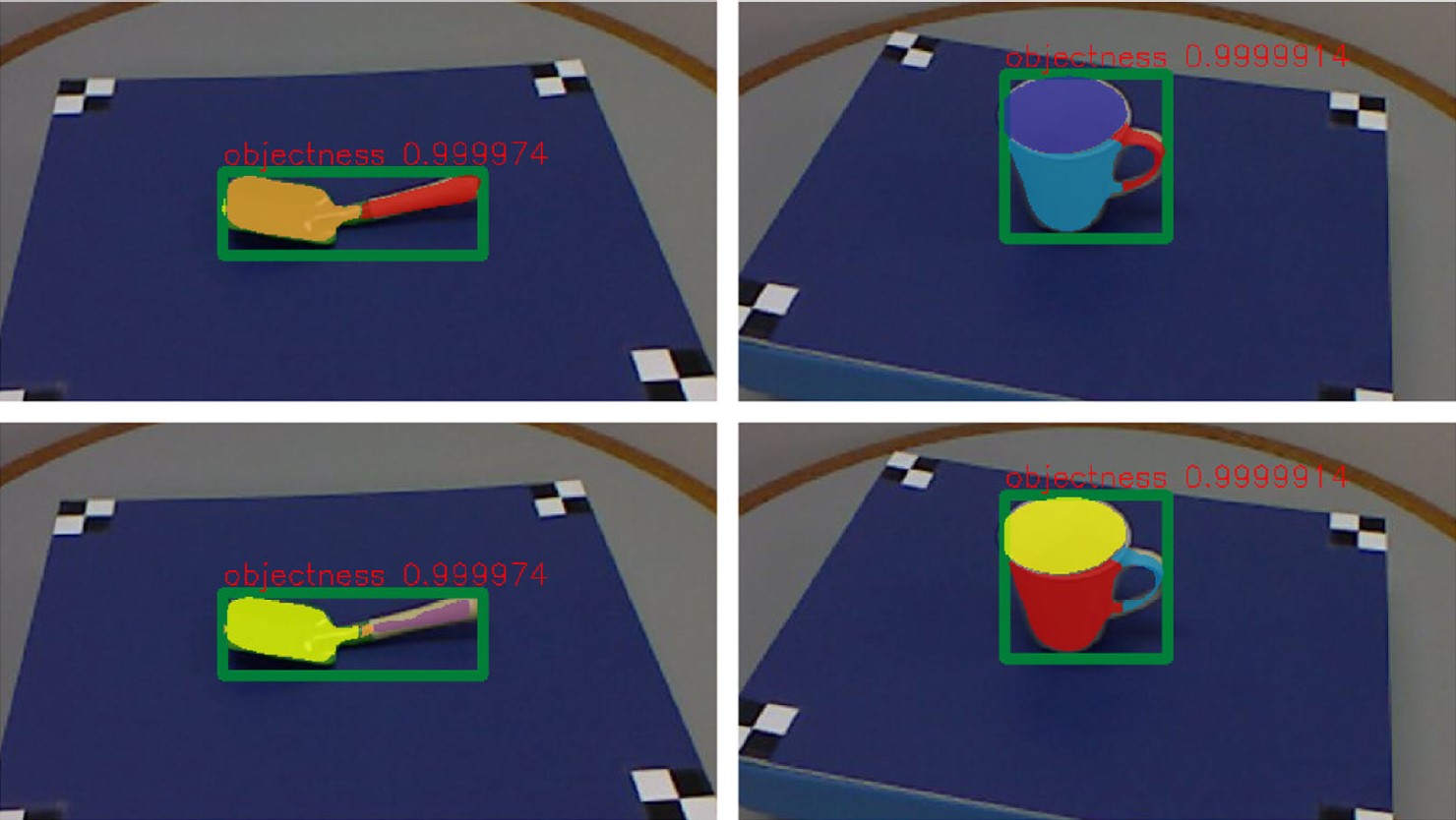}
	\caption{Segmented image from \cite{chu2019toward}. Red parts afford grasping, orange afford supporting, deep blue afford containing, blue afford wrap-grasping, and purple afford pounding. \label{fig:sug-cup}}
\end{figure}

Many works propose to model what actions are afforded on which part of an object as an image or point cloud segmentation problem \cite{myers2015affordance, kokic2017affordance, nguyen2017object, do2018affordancenet, chu2019toward, mandikal2020graff, hamalainen2019affordance, borja2022affordance}. In these works, a segmented part of an object image or point cloud is labelled with one or more affordable actions, i.e., a binary mask that indicates whether an action can be applied to that part of the object. The action possibilities are simplified into binary variables and represented as pixel-level masks. For example, as shown in Fig.~\ref{fig:sug-cup}, the pixels or points of the handle of a cup are labelled as being graspable, while those of the hollow part of the cup are labelled as containable. It is common for different parts of an object to have different affordances. It is also common for the same part of an object to have multiple affordances \cite{chu2019toward}. 

As a natural extension, these pixel-level or point-level affordance predictions were used to provide the downstream manipulation policy with extra task information. The most straightforward way in grasping tasks is to designate the centre of the detected affordance masks as a grasping location \cite{chu2019toward}. A more recent method treated the predicted segmentation masks as an extra channel of the image observations. A manipulation policy then processed this extended image to determine what actions to take \cite{mandikal2020graff}. A self-supervised learning method was proposed to learn to predict the pixel masks for gripper-object interaction centres from human teleoperation demonstrations of a table tidy-up task \cite{borja2022affordance}. These pixel masks were then used in the real world for a model-based policy to move the gripper closer to the interaction point of an object and a reinforcement learning policy to pick up the object. There was also an attempt to learn a latent representation of object affordances with Variational Auto-Encoders \cite{Kingma2014, hamalainen2019affordance}. It was successfully trained using simulation data and transferred to a real-world robotic system, aided by domain randomisation technique. They used the latent representation to generate robot trajectories that move the gripper to a point above a cup \cite{hamalainen2019affordance}.

\subsection{Action scores}\label{subsec:AS}

Several works proposed to represent the action possibility as a continuous variable that indicates how confident it is that an action can be successfully executed (is affordable) \cite{zeng2018robotic, cai2019metagrasp, wu2020learning, yang2021learning, mo2021where2act, ichter2022do}, while the segmentation masks discussed in the last subsection are binary variables.

\begin{figure}[t]
    \centering
	\subfloat[Action score prediction for four kinds of primitives \cite{zeng2018robotic}.]{\includegraphics[width=0.48\textwidth]{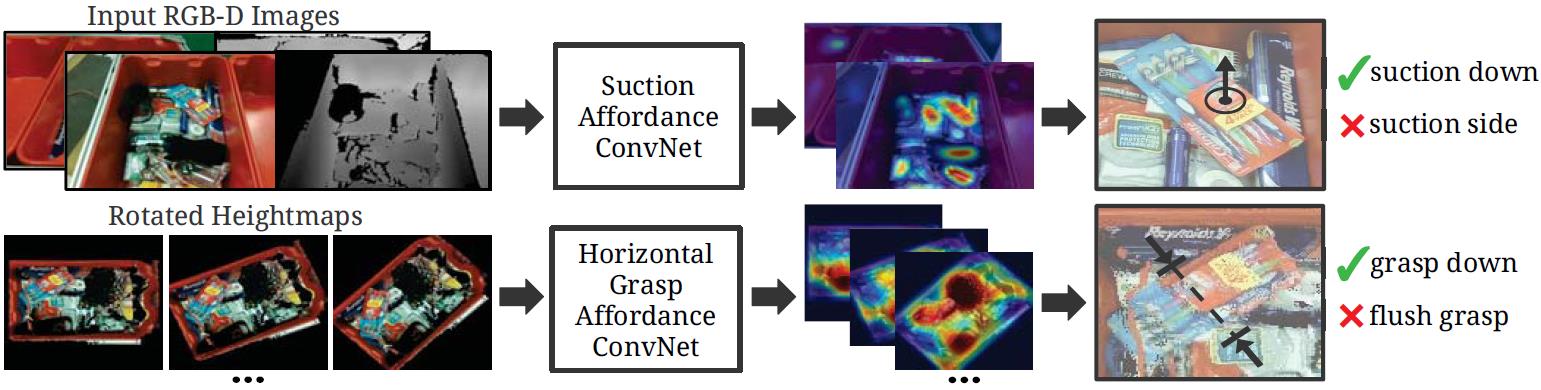}}\\
	\subfloat[Grasping success score prediction \cite{wu2020learning}. ]{\includegraphics[width=0.48\textwidth]{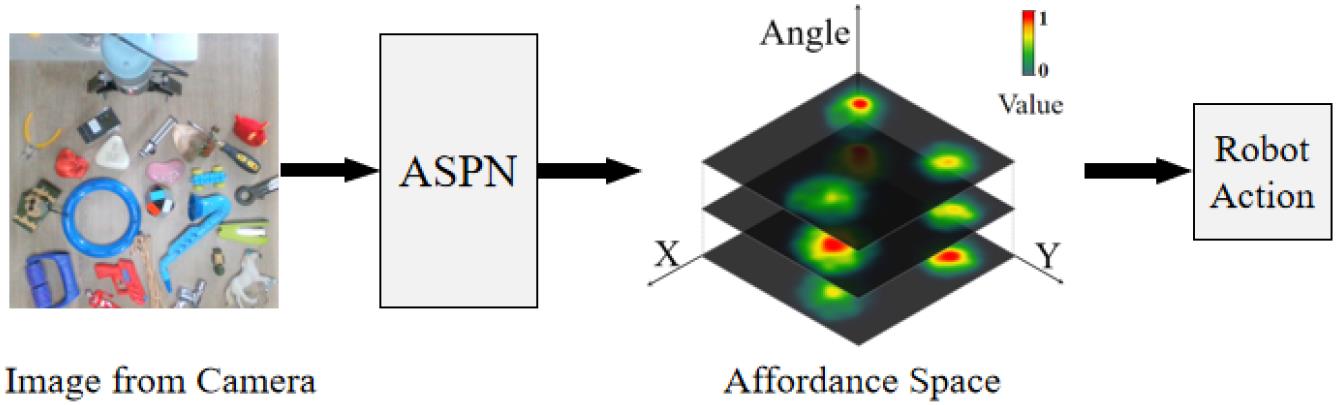}}
	\caption{Examples of action score prediction. \label{fig:action-score}}
\end{figure}

\citeauthor*{zeng2018robotic} proposed to model the success probabilities of four kinds of primitive grasping and suction actions given the RGB-D observation of a clutter scene \cite{zeng2018robotic}. The probability distributions are defined as matrices whose entries represent the success rates of executing actions at the pixel locations (see Fig.~\ref{fig:action-score}). Similarly, \citeauthor*{cai2019metagrasp} proposed to predict graspability, ungraspability and background affordances over image pixels, achieving a grasping success rate of $93\%$ on a set of household items, $91\%$ on a set of adversarial items and $87\%$ in clutter scenarios \cite{cai2019metagrasp}. The network was trained with synthetic data generated by an antipodal grasp heuristics in simulation in a self-supervised fashion. \citeauthor*{wu2020learning} extended such a 2D affordance map defined in the pixel space into a 3D space, estimating the graspability not only in different x-y positions, but also in different grasping angles \cite{wu2020learning}. Another work proposed to first train a neural network to predict object classes and segmentation masks of a clutter scene, and then train a DQN network to predict the grasping success scores based only on the segmentation masks \cite{yang2021learning}. This work successfully transferred the learnt grasping score prediction system to the real world with domain randomisation. Recently, \citeauthor*{mo2021where2act} proposed to predict action scores for a set of six motion primitives based on RGBD images or point clouds. They designed a three-branch network architecture to 1) predict the actionability of a pixel or a point, 2) propose gripper orientations and 3) estimate the success score of the primitive action given the action pixel and orientation \cite{mo2021where2act}. In another interesting recent work \cite{ichter2022do}, the authors propose to represent the action possibilities of a large number of pretrained motion skills by the action value function in the RL framework based on RGB observations. These papers are closely related to the works in vision-based robotic grasping (VBRG), where many works were not linked to the concept of affordance. For a thorough review for VBRG, please refer to \cite{du2021vision, marwan2021comprehensive}.

\subsection{Summary and limitations}\label{subsec:MAP-Limit}
To summarise, though some recent works tried to estimate action possibilities for a variety of actions, most of them focused on grasping tasks when deploying the learning system. These works leveraged motions that are generated by a motion planner or hand-crafted by humans. In terms of affordance learning, they sought to estimate whether a planned motion or primitive can be successfully performed at an image pixel location or a point in the point cloud. The learnt affordance model was used to infer a desired action by extracting a pixel location or a point that is centred at the affordable region or with the highest action possibility. There are several limitations regarding the papers discussed in this section.

1) At the current research stage, the community lacks an image segmentation dataset for object affodances at large scale \cite{ardon2021building}, when compared to datasets like COCO \cite{lin2014microsoft} or ImageNet \cite{deng2009imagenet}. It is promising to build larger datasets, as demonstrated by the ImageNet dataset for image classification, though a vast amount of human labour is required. To reduce such human labour, self-supervised learning techniques could be employed, such as automatic labelling \cite{chen2020big, borja2022affordance} and interactive labelling \cite{chegin2020interactive}.

2) Though multi-affordance detection has drawn researchers' attention \cite{do2018affordancenet, chu2019toward}, real-world manipulation experiments using affordances are restricted to only one or two categories (mostly grasping) \cite{kokic2017affordance, nguyen2017object, do2018affordancenet, hamalainen2019affordance, chu2019toward, mandikal2020graff, borja2022affordance, zeng2018robotic, cai2019metagrasp, yang2021learning}. Not much attention was given to other actions such as push and pull \cite{wu2020learning, mo2021where2act}. In addition, they are subject to fully or partially hand-crafted motion primitives (e.g., top-down parallel-jaw grasping), thus are limited to a very small set of object-action relationships. For example, they cannot represent affordances for 6DoF grasping actions or non-primitive interactions. A recent work in coupling language instructions and mobile robot motion skills makes a pioneering example on more complex action affordances learning and real-world grounding \cite{ichter2022do}.

3) These methods only predict action possibilities, ignoring the knowledge about the effects of these actions. From a human perspective, we tend to use affordance knowledge for planning, which requires us to be aware of not only what the possible actions are, but also what the results of these actions are. The next section elaborates on recent attempts to incorporate both action possibilities and effects.

4) These works exclude the dependencies between the executions of multiple actions and the influences of different manipulation objectives. For example, the possibilities of grasping a cup at its handle would differ when the robot is tasked to hang it up, place it on a table or hand it out to another agent. This involves a planning process for different final task objectives. We discuss more on this point in the next section.

\section{Keypoint Affordance}\label{sec:KA}

\begin{figure}[b]
    \centering
	\includegraphics[width=0.48\textwidth]{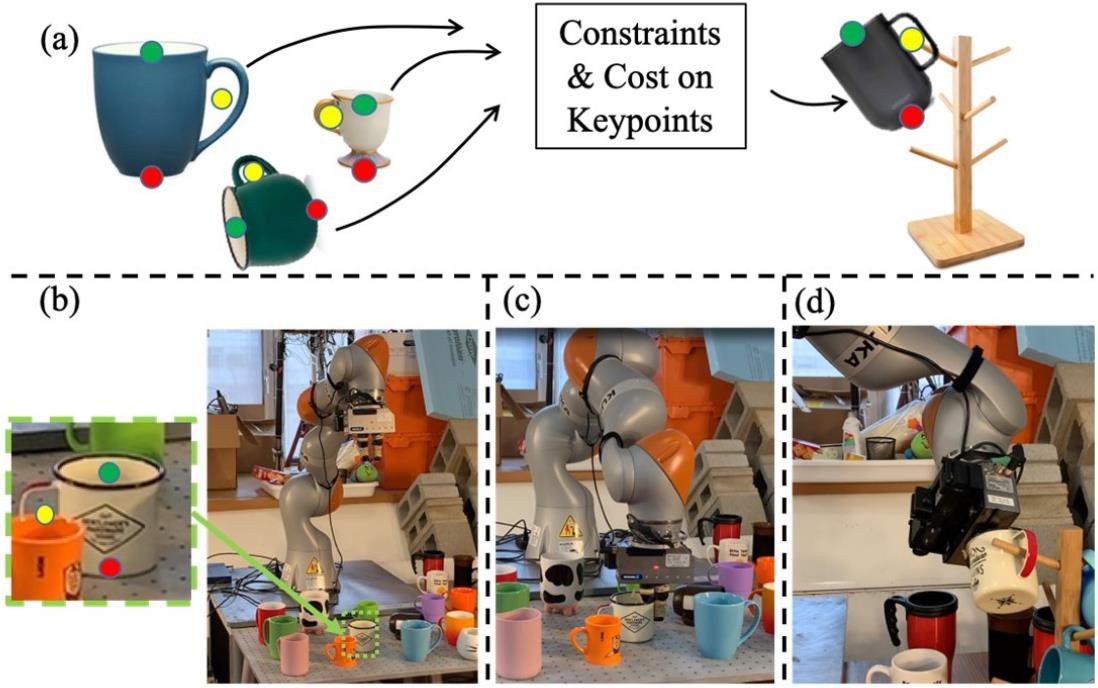}
	\caption{Category-level keypoint detection from \cite{manuelli2019kpam}. (a) Detected keypoints for different cups in planning; (b) keypoint detection; (c) grasping; (d) hanging. \label{fig:kpam}}
\end{figure}

In the last section, we discuss papers that sought to first compute the action possibilities, $\hat{p}^{\mathcal{AF}}$, and then infer the afforded actions from the action possibilities. For example, compute a binary or continuous matrix that indicates whether a gripper can pick up an object at each pixel location of an RGBD image. In these cases, a pixel in an image or a point in a point cloud is associated to an action as a parameter of a motion planner or a primitive. 

In this section, we review works that proposed to generate the afforded actions by predicting object keypoints, skipping the computation process of the action possibility \cite{manuelli2019kpam, gao2021kpam2, gao2021kpamsc, qin2020keto, turpin2021gift, xu2021affordance}. The keypoints were defined as the functional points of an object. They were associated with affordance because they could be used by some action inference methods (e.g., a motion planner) to generate afforded actions. Keypoints provide the action inference method with a smaller search space and easier-to-define task-relevant geometric constraints. From the RL perspective, the keypoints can be seen as an abstract observation that indicates the action space for a policy or value function, or itself as a constrained action space that corresponds to a set of affordable motion primitives. The later one is adopted by many previous works. Previously, keypoint methods with non-deep learning techniques were limited to specific objects of a particular shape and size \cite{jamone2016affordances}. In this review we focus on deep learning-based methods that are able to generalise to unseen and novel objects \cite{qin2020keto, manuelli2019kpam, gao2021kpamsc, gao2021kpam2, xu2021affordance, turpin2021gift}. A summary of the observations, object types, training methods, deployment tasks and motion generation methods of these works are given in TABLE~\ref{tab:keypoint-summary}.

\begin{table*}[t]
\caption{Summary of papers focused on affordance keypoint prediction. \textbf{PCD:} point cloud data; \textbf{SL:} supervised learning; \textbf{SSL:} self-supervised learning; \textbf{Sim:} simulation \textbf{Real:} real-world; \textbf{DoF:} degree of freedom; \textbf{PJG:} parallel-jaw grasp; \textbf{MPC:} model predictive control.}    
\label{tab:keypoint-summary}
\begin{center}
\begin{tabular}{c|c|c|c|c|c|c}
\hline
Paper                        & Object classes   & Affordance (afforded actions) & Input & Method & Deployment Task & Motion\\
\hline
\cite{manuelli2019kpam} & Shoes; Mugs    & 6DoF PJG, place \& hang & RGBD & SL & Shoe-placing, mug-placing \& mug-hanging (Real) & Planning\\
\cite{gao2021kpam2}     & Erasers; Pegs; Holes & 6DoF PJG, wipe, insert & RGBD & SL & Whiteborad wiping, peg-in-hole insertion (Real) & Planning\\
\cite{gao2021kpamsc}    & Shoes; Mugs    & 6DoF PJG, place \& hang & RGBD & SL & Same as \cite{manuelli2019kpam} with shape completion (Real) & Planning\\
\cite{qin2020keto}      & Hammers        & 6DoF PJG, hammer, push, reach & PCD & SSL & Object hammering, pushing \& reaching (Sim) & Planning\\
\cite{turpin2021gift}   & Hammers        & 4DoF PJG, hammer, push, hook & RGBD & SSL & Object hooking, reaching, hammering (Sim) & MPC\\
\cite{xu2021affordance} & UMD+GT dataset & UMD+GT dataset & RG-D & SL & PJG, pouring, arranging, cutting (Sim \& Real) & Planning\\
\hline
\end{tabular}
\end{center}
\end{table*}

\citeauthor*{manuelli2019kpam} proposed kPAM, which defined keypoints for objects that belong to the same category (Fig.~\ref{fig:kpam}) and supported grasping, placing and hanging actions to be inferred from the keypoints. For example, three keypoints at the handle, top and bottom for mugs. These keypoints were predicted given a segmented RGBD image and then used by a motion planner to generate motions for pick and place tasks. The authors later formulated a feedback control framework with keypoint-based object and action representations, and accomplished a peg-in-hole insertion task with a variety of objects \cite{gao2021kpam2}. They also extended the method to include a shape completion technique, named kPAM-SC, so that the generated motions can handle object collision \cite{gao2021kpamsc}. Another work, KETO, used a three-keypoint pattern, including a grasp point, a function point and an effect point, to represent hammer-like tools and infer hammering motions \cite{qin2020keto}. A generative network was trained to produce keypoint candidates given an object point cloud. An evaluation network was trained to predict the manipulation success scores for these keypoints. The training process was conducted in a self-supervised manner using task completion signals. These keypoints, along with a set of task keypoints within a simulation environment, were used to generate motions by solving a Quadratic Programming problem \cite{qin2020keto}. \citeauthor*{turpin2021gift}, proposed GIFT, which predicted a set of representational keypoints for an object and then selected from them a grasping point and an interaction point. This procedure allowed the functional keypoint pattern to be discovered instead of being specified by users. They represented the functional keypoint proposal model as a Graph Neural Network (GNN) over the representational keypoints. They then computed a robot motion using model predictive control and evaluated the task-specific return for the motion. The functional keypoint proposal model was trained by optimising an REINFORCE loss with the task-specific return. 

Instead of predicting keypoints for a category of objects as done in \cite{manuelli2019kpam, gao2021kpam2, gao2021kpamsc, qin2020keto, turpin2021gift}, \citeauthor*{xu2021affordance} proposed to define keypoints for afforded actions on images \cite{xu2021affordance}. They modified the affordance image segmentation dataset UMD \cite{myers2015affordance} by assigning a set of five 2D keypoints to each affordance region. These keypoints defined the position and direction information about the afforded actions. They proposed a two-branch deep neural network, AffKp, to learn affordance image segmentation and keypoint detection in parallel via supervised learning. The predicted keypoints were projected from the image plane to the real-world frame and used to infer the corresponding afforded actions.

\textbf{Summary:} To sum up, these works proposed to infer afforded actions that manipulate an object from a set of keypoints defined on the object. According to the affordance definition introduced in section~\ref{sec:ADMDP}, they are classified as methods that compute the afforded actions, rather than compute the action possibilities. For example, to infer various grasping configurations from a predicted grasping point on a tool handle \cite{qin2020keto} instead of a set of action possibilities \cite{zeng2018robotic}. Most of the works leveraged human knowledge to create a pattern of keypoints and trained deep neural networks to predict them for a category of objects \cite{manuelli2019kpam, gao2021kpam2, gao2021kpamsc, qin2020keto, xu2021affordance}, while only one work, GIFT, proposed to discover functional keypoints using task-completion signals \cite{turpin2021gift}. The main benefits of using keypoints to infer afforded actions include but not limit to: 

\begin{itemize}
    \item keypoints can capture the common properties of a category of objects;
    \item keypoints can support the inference of various afforded actions;
    \item keypoints can be used to reduce the searching space of afforded actions for the action inference processes.
\end{itemize}

\textbf{Limitations:} The primary limitation of keypoint-based methods is that pre-defining a fixed pattern of keypoints requires a relatively large amount of human prior. This eases the keypoint prediction model from the difficulty of learning from scratch, but limits the generalisability of the learnt keypoint patterns. In reality, one specific pattern of keypoints is unlikely to be sufficient and flexible enough for the diverse manipulation tasks that may need to be performed on the objects. The aforementioned papers have evaluated their methods on tasks with relatively simplified geometric constraints and manipulation skills \cite{qin2020keto, manuelli2019kpam, gao2021kpamsc, gao2021kpam2, xu2021affordance}. For example, when a robot could only reach a hammer's head, it could not grasp the head and use the handle as a hammering point if it can only recognise the head as a hammering point. Learning to predict keypoint patterns with free interactions and task-completion signals is promising for reducing such human biases \cite{turpin2021gift}.

Secondly, sparse keypoint representation is not very compatible for tasks that are sensitive to object shapes and sizes, when compared to a full point cloud representation. For example, when manipulating a deformable object like a soft plastic cup, keypoints are not enough for the robot to determine the grasping force and track the deformation of the cup \cite{arriola2020deformablereview}. In this regard, multi-modal representations may be required, such as using keypoints along with a shape-completion procedure \cite{gao2021kpamsc}. In the future, other observation modalities, such as tactile sensors, force sensors, etc., may be incorporated with keypoints to better infer afforded actions in real-world manipulation tasks.

Last but not least, the primary method to infer afforded actions using keypoints, motion planning, is difficult and expensive in environments with complex dynamics and large action and state spaces. It poses two problems to classic methods: 1) user-specified dynamic models have difficulties to represent highly stochastic and non-linear real-world systems and to generalise to high-dimension inputs like images and 2) planning over large action and state spaces is very expensive and difficult. Researchers have proposed to address them by learning a system dynamic model from data \cite{polydoros2017survey, hafner2019learning, eiffert2020path, qureshi2020motion, moerland2020model}, though they did not explicitly consider the concept of affordances. We elaborate in the next subsection on recent works that propose to plan robot motions using a learnt affordance-aware dynamic model.

\section{Modelling Action Possibilities and Effects}\label{sec:MAPE}

As defined in section~\ref{sec:ADMDP}, the effects of afforded actions can be modelled by a partial dynamic model $\hat{P}_I(s'|s,a)$, which predicts the next system states given a pair of state and \textit{afforded} action. The motivation of building a dynamic model is to equip a robot with a safer and more efficient method to generate motion plans or learn from imagined data. A dynamic model releases the robot from expensive and potentially unsafe interactions with the real world \cite{polydoros2017survey, moerland2020model}. Previous works on action effect modelling have relied extensively on manually-abstracted state representations and dynamics \cite{csahin2007afford, ugur2015bottom}, which has a deep connection to the field of symbolic planning \cite{karpas2020automated}. It is difficult, however, to hand-craft dynamic models for real-world systems with complex observations. Therefore, in recent years researchers have proposed deep learning methods to learn the dynamic model from data, demonstrating the value of having access to a dynamic model over the space of complex sensory observations \cite{finn2017deep, polydoros2017survey, moerland2020model}.

\begin{figure}[h]
    \centering
	\includegraphics[width=0.48\textwidth]{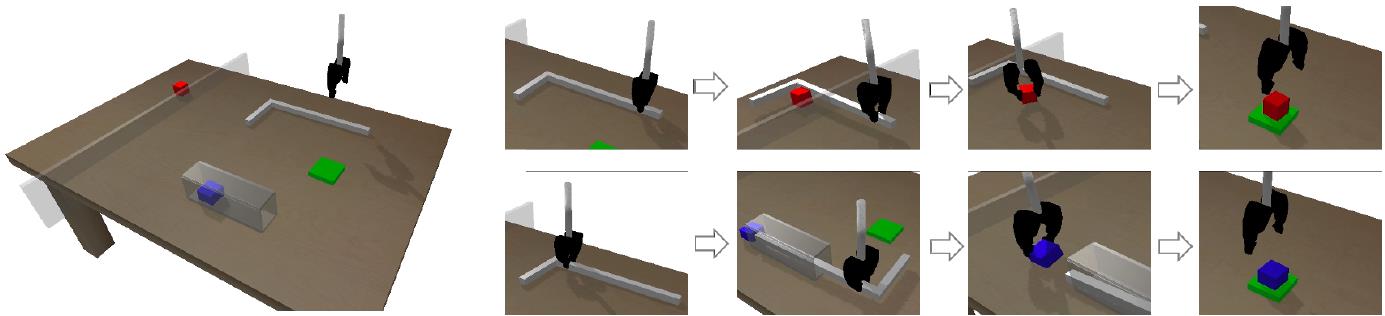}
	\caption{The multi-step tool-use task designed to evaluate the Deep Affordance Foresight method proposed in \cite{xu2020daf}. The robot needs to decide which end of the L-shape stick to grasp for reaching the red block or push the blue block out of the tube. \label{fig:daf}}
\end{figure}

Among many recent advances of learnt world models, \citeauthor*{khetarpal2020can} proposed to integrate the concept of affordances in the model-based reinforcement learning (MRL) paradigm (as rephrased in section~\ref{sec:ADMDP}). They first learnt a binary classification model to predict whether some actions are afforded given an observation, which was essentially estimating the action possibilities $\hat{p}^{\mathcal{AF}}$ as binary variables. Different from methods discussed in section~\ref{sec:MAP}, they did not infer the afforded actions from the estimated action possibilities. Rather, they proceeded to learn a dynamic model of the world for only actions that were classified as possible or effective. Data of non-effective actions are regarded as redundant and ignored. The resultant model was a partial dynamic model (PDM) of the system. During planning, the PDM is only queried for effective actions according to $\hat{p}^{\mathcal{AF}}$. In short, the benefits of such a framework are twofold: 1) it accelerates planning by only considering the afforded actions and 2) it accelerates dynamic model learning by focusing on learning part of the system dynamics concerning the afforded actions of interests. They were demonstrated first in a continuous 2D navigation task in \cite{khetarpal2020can} and later in unseen long horizon manipulation tasks in simulation with image inputs (Fig.~\ref{fig:daf}) \cite{xu2020daf}. This affordance-aware model-based reinforcement learning framework was later extended to develop temporally abstract partial dynamic models, considering options (sub-policies) that are only afforded in certain situations. The authors empirically demonstrated the success of learning option affordances and partial option models online, resulting in more efficient learning and planning in a 2D Taxi task \cite{khetarpal2021temporally}.

\textbf{Limitations:} As a relatively new direction, the first limitation is the lack of evaluation in more realistic examples. Most previous works are performed in simulation using synthetic data. Tasks with image or point cloud observations from real robots with longer time horizon would increase the complexity considerably. More efforts are required to design more realistic tasks.

Secondly, the predicted action effects in the proposed examples are more short-term or instant effects of single-step action commands. In practice, planning is often more valuable with macro actions that consist of a series of single-step control commands, exhibiting a particular kind of skill, such as pushing for a certain distance, approaching and grasping an object, lifting up for a certain height, etc. This requires the algorithm to reason about long-term action possibilities and consequences. Though an attempt was made to incorporate affordances with temporally abstract partial models for more efficient planning at a more abstract level, it was only evaluated in a 2D Taxi task \cite{khetarpal2021temporally}. More effort is needed to evaluate and improve its performance on robotic tasks in the future.

Thirdly, the proposed method focuses on affordances of a given state, which is likely to be computationally inefficient for tasks with complex observations containing diverse information irrelevant to the manipulation goal. From a human perspective, we typically only attend to some parts of the observation that are most relevant to the task of interest, saving energy and improving planning efficiency and accuracy.

\section{Discussions and challenges}\label{sec:discussion}

According to the reviewed papers, this section summarises the limitations of deep robotic affordance learning (DRAL) and identifies its bottlenecks at the current stage. We conduct the discussion and pose future research challenges from the following angles: observations, actions, affordance representations, data collection and real-world deployment.

\subsection{Observation} 
For most tasks, especially real world tasks, a robot relies on sensors to perceive the environment \textit{without the access to the true system dynamics} such as the velocities of objects. This is one of the most common assumptions adopted by robotic researchers. Previous works have spent efforts developing symbolic representations for the observations of the system to simplify the mapping from sensory observations to affordances \cite{yamanobe2017brief, cruz2018multi}. In recent DRAL literature, the types of observations have become more complex, inluding object states (normally in simulation), object point clouds and RGB(D) images. 

Another important assumption made by these works is that \textit{the observation contains enough information to reason about affordance}. However, this does not always hold true. For example, a heated plate may be detected as graspable from RGBD or point cloud observations though it may be actually too hot to hold by a human. Some affordances may require information about temperature, softness, transparent surface, reflection, etc., that are difficult for (depth) cameras to capture. It is also worth-noting that languages are becoming more popular to provide instruction or extra information about the desired tasks and skills for affordance learning \cite{jiang2019language, ichter2022do} due to the rise of large language model (LLM). Information about the robot itself, such as sensorimotor states, could also help to reason about affordances like reachablility. On the other hand, affordances of occluded objects are difficult to detect from a fixed camera view point. Combining all these, a promising direction for future research is to apply multi-modal and multi-viewpoint observations for affordance detection \cite{cruz2018multi, lee2020making}.

The third assumption about observations, especially for deep learning-based methods, is that \textit{the mapping from inputs to actions or action possibilities can be found through gradient descent}. However, given the large space of observations in the real world, it is very challenging to find such a mapping even it does exist. Some works applied pre-processing methods to help the robot focus on the most relevant information for affordance learning or action inference, such as applying object masks \cite{yang2021learning} or extracting object keypoints \cite{manuelli2019kpam}. Such ideas make computation more efficient by shrinking the size of observation space, whereas more or less lose some degree of generality due to human priors. In this regard, future research could focus on representation design or learning, giving special attention to the trade-off between generalisability and learning efficiency (or computational cost) for affordance detection or afforded action inference. 

\subsection{Action}

Noticeably, researchers preferred motion primitives in recent DRAL works. For example, grasping primitives that move a gripper towards an identified grasping location and close the fingers \cite{wu2020learning, yang2021learning}, and placing primitives that move a gripper with an object to a location and release the fingers \cite{manuelli2019kpam, gao2021kpamsc}. Note that these primitives can be motions planned by a planner \cite{qin2020keto, wu2020learning, manuelli2019kpam, gao2021kpamsc} or parameterised motor skills \cite{xu2020daf}. These primitives exhibit relatively simple motions, such as pick-and-lift \cite{yang2021learning, mandikal2020graff, chu2019toward, wu2020learning, cai2019metagrasp}, pick-and-place \cite{gao2021kpamsc, manuelli2019kpam, zeng2018robotic}, pushing \cite{xu2020daf} and hammering \cite{turpin2021gift, xu2021affordance, qin2020keto}. The use of motion primitives as actions exhibits a trend that the community is more interested in the affordances of high-level skills, rather than low-level control commands. To follow this trend, we pose some challenges and future directions to consider. 

The adaptability of the primitive motions considered by recent works could be improved, as they were mostly designed for open-loop control. For example, given a grasping point, a grasping motion moves the gripper to the grasping point and closes the fingers, without any adaptation in between. However, the detected grasping affordance may be inaccurate or changed during the execution of the motion due to occlusion, human factors, collision with the robot arm or finger slippery, etc. To cope with such challenges, one may consider a feedback control style method for action inference \cite{gao2021kpam2, turpin2021gift}. Another interesting direction to consider is an algorithm that is permitted to stop and re-select motion primitives. For example, when an insertion motion changes from affordable to unaffordable, the robot may select a re-position motion without waiting for the insertion motion to reach its execution time limit. The notion of \textit{interrupted options} based on the option framework \cite{sutton1999between} may serve as a good theoretic foundation.

Predefined primitive motions are very useful when the manipulation task is in a rather structured environment without unexpected factors. However, the real world is highly unstructured and uncertain. A robot needs to generalise its skills to novel situations quickly or sometimes finds new skills to manipulate an object. This means the robot may be required to discover new afforded actions. To achieve this, the action space needs to be general enough. One promising direction is the study of option or subgoal discovery in hierarchical reinforcement learning \cite{wulfmeier2021data}, in which skills (in the form of sub-policies) are discovered instead of predefined.

\subsection{Affordance Representations}\label{subsec:Reprt}
According to Gibson \cite{gibson1982concept}, perceiving affordance does not need information processing or any internal representations, but only requires the extraction of fundamental physical properties of the target object or environment. For example, perceiving that a needle has a pointed end leads to the perception that the needle affords piercing. This reasoning is theoretically sound \cite{brooks1991intelligence} but is however practically limited as in practice, some form of mathematical representation of affordances is required to facilitate action inferences \cite{hassan2021affordance}. Also, it is important to note that there is so far no known widely-adopted benchmarking metrics for qualitative or quantitative comparative studies of different representations proposed in the field. What intermediate representations are needed in the spectrum between end-to-end learning and manually constructing everything is mostly specific to the problem of interest.

As this review is inclined to the recent practical applications of affordances in DL-powered RL and robotics, is seems more graspable and plausible from a practical standpoint to discuss the representations of affordances in recent literature according to \textit{how the action inference method works}. Afforded actions are inferred in mainly three manners: 1) from the action possibility estimates, 2) by a direct mapping from the observations and 3) by planning with a partial dynamic model. The first and third classes require an explicit representation of the action possibilities and effects, while the second one may need an intermediate representation that constrains the action space (such as object keypoints).

Action possibilities for primitive motions were represented often by an \textit{affordance map}, which is typically a matrix that has the same size of the observation image. Its entries indicate the success rates or possibilities of executing certain primitive motions at the corresponding pixel locations \cite{wu2020learning, cai2019metagrasp, zeng2018robotic, yang2021learning}. Segmentation masks can be regarded as a special case with binary variables \cite{myers2015affordance, mandikal2020graff, chu2019toward, hamalainen2019affordance, do2018affordancenet}. It can also be applied to point clouds in the 3D space \cite{kokic2017affordance, mo2021where2act}. This representation is efficient as it estimates the possibilities for a set of actions simultaneously, but is limited to primitive motions that operate over the discrete image pixels or object points. It may not easily generalise to continuous observations such as sensorimotor states, force feedback, etc. For actions that are not parameterised on images or point clouds, one may need to represent the action possibilities as a classifier \cite{khetarpal2020can}. In order to scale to real-world tasks, it is promising to develop methods to accelerate the learning of the action possibility estimator with large and continuous action space, such as learning from demonstrations \cite{nair2018overcoming}.

Representing and predicting the effects of actions is another difficult topic. Though an action possibility estimator helps to reduce the learning data requirement and increase the planning efficiency for dynamic models \cite{khetarpal2020can, xu2020daf}, the difficulty of reconstructing high dimensional observations (e.g., image or point cloud) remains. Experiences and methods from other fields could be considered, such as video prediction \cite{wu2021greedy}. There is also a large body of works devoted to the learning of dynamic models \cite{moerland2020model}. Abstract representation for system observations is another closely related topic \cite{abel2019theory}. Future research may focus on applying general dynamic model learning methods to partial dynamic models with an action possibility estimator. Another challenge in the long term may be how the learning of affordances affect the learnt representation of the world, which is related to the topic of understanding the world through interaction.

Another way to compute afforded actions in the literature is through a direct mapping from observations to a set of afforded actions. The crucial question is how to represent the scene/object in a way that relate to their afforded actions. One popular solution is to use object keypoints that geometrically capture some functions of a category of objects, such as grasping points of mugs \cite{qin2020keto, manuelli2019kpam, gao2021kpamsc, gao2021kpam2, xu2021affordance, turpin2021gift}, as discussed in section~\ref{sec:KA}. From the keypoint methods we can identify some criteria to be satisfied when considering other types of representations. These include: 1) intuitive or convenient for generating robot motions; 2) able to generalise cross robot hardware (grippers, arms, etc.); 3) able to capture the common properties of many objects. Notice that such a representation should be designed as an abstraction of the observations of a scene or an object that relates to the afforded actions. The keypoint-based methods rely on motion planning or model predictive control to generate the desired motions (see Table~\ref{tab:keypoint-summary}), while one may come out with representations that suit other motion generation techniques (e.g., reinforcement learning, imitation learning, etc.).

\subsection{Data collection}\label{subsec:DC}

Deep learning methods require a considerable amount of data to achieve good generalisation performances \cite{goodfellow2016deep}. Previous papers in DRAL have used supervised learning, self-supervised learning and reinforcement learning as their core training methods, each of which has a unique data collection process. 

Supervised learning methods rely fully on human prior to collect and generate data, which is expensive for large datasets (e.g., ImageNet \cite{deng2009imagenet}). Most papers use the UMD dataset \cite{myers2015affordance} for evaluation. However, it only provides segmentation labels. To alleviate the difficulty of collecting manipulation-specific data (e.g., grasping points, motion trajectories, etc.), some papers adopt self-supervised learning to collect data automatically through simulations \cite{yang2021learning, wu2020learning, cai2019metagrasp, qin2020keto, turpin2021gift}. Reinforcement learning (RL)-based methods generate training data by interacting with the environment using a learnt policy with some degree of randomness \cite{sutton2018reinforcement}. In addition, the performance of the RL policy is evaluated directly on task return or success rate, without intermediate metrics (e.g., accuracy of predicting segmentation masks or keypoints). However, off-policy RL methods can benefit from data generated from other sources, such as human demonstrations \cite{nair2018overcoming}.

A limitation, at the current stage, is the lack of a consensus on which benchmark should be used to generate the data and evaluate the algorithms for DRAL. Ideally, such a benchmark should provide handy Application Programming Interfaces (APIs) and functions to support the data collection processes for supervised, self-supervised and reinforcement learning. Common functionalities, such as capturing RGB(D) images and point clouds, classic planning algorithms, popular RL baselines, etc. are also considered helpful. It could be more valuable if tasks that feature multiple manipulation objectives and multi-step manipulation are designed and built-in. There are several open-source datasets, simulation environments or benchmarks that may be extended for such purposes \cite{james2020rlbench, batra2020rearrangement, liang2018gpu, mo2021where2act}. The community has not yet seen a large scale dataset for DRAL that covers the mentioned aspects.

\subsection{Real-world deployment}\label{subsec:RwDep}

For methods that use real-world data, the main difficulty is primarily the expensive data-collection process, which was covered in the last subsection. The main concern that arises during the final deployment or evaluation is then the insufficient generalisation ability, which is largely caused by the limited amount of training data. 

1) \textit{Supervised learning} methods are easier to be deployed in the real world after being trained, though their performances rely extensively on the quality of the dataset. In the past few years, many datasets that support the learning of stable grasping have been constructed \cite{fang2020graspnet, morrison2020egad, myers2015affordance, zeng2018robotic, du2021vision, marwan2021comprehensive}. However, very few are built for multiple manipulation objectives or multi-step tasks \cite{huang2019dataset, mandlekar2019scaling}. Consequently, more efforts are needed to collect data that cover diverse background textures, view-points, objects (in terms of types, shapes, dimensions, etc), manipulation skills (trajectories) in order for supervised learning-based DRAL to work in the real world. 

2) \textit{Reinforcement learning} in the real world is even more difficult due to the high risk of hardware damages during exploration and a considerable amount of human labours for resetting the environment \cite{ibarz2021train}. 

3) \textit{Sim2real transfer} is another stepping stone for successful real-world deployment, as researchers have resolved to training in simulation to avoid the painful data-collection process in the real world. Inevitably, deploying models trained in simulation onto the real-world systems will have to face the simulation-to-reality gap. In order to cope with such differences, researchers have proposed to use domain randomisation to extend the distribution of training data \cite{tobin2017sim2real}. It can be applied to image textures \cite{yang2021learning, hamalainen2019affordance, tobin2017sim2real, james2019sim}, camera parameters \cite{ren2019domain} and physical properties \cite{ibarz2021train}. Recent DRAL works limit their real-world applications within a relatively unchanged and structured environment. Long horizon tasks that require the reasoning of the long-term effects of diverse skills or objects have mainly been studied in simulation. More efforts are needed to evaluate and adapt existing methods to real world data.

\section{Conclusion}\label{sec:Concl}

This review paper looks into the recent advances in the topic of deep robotic affordance learning (DRAL). DRAL aims to develop data-driven (deep learning) approaches to apply the concept of affordance to robotic tasks. We suggest in this review to summarise and analyse these works based on the reinforcement learning (RL)-based definition of affordances \cite{khetarpal2020can}. We briefly recall this definition in Section~\ref{sec:ADMDP}, where we classify recent DRAL papers and discuss the connections between RL and affordances. Accordingly, they are categorised into three classes of works that: 
\begin{itemize}
    \item 1) infer afforded actions from the estimated action possibilities;
    \item 2) learn an abstract object/scene representation that relates to the set of afforded actions;
    \item 3) generate afforded actions through planning with a learnt partial dynamic model and an action possibility classifier.
\end{itemize}
Advances and limitations of the three lines of works are discussed in section~\ref{sec:MAP}, \ref{sec:KA} and \ref{sec:MAPE}, respectively. A more general discussion for the field and its challenges are given in section~\ref{sec:discussion}.

\textbf{Final remark:} We further propose here a promising direction to extend the RL-based affordance definition. In \cite{khetarpal2020can}, the intent captures the desired resultant state of an action taken at a system state. Subsequently, the corresponding affordance is defined as a subset of state and action pairs in which the intent is satisfied. In \cite{khetarpal2021temporally}, the definitions of intent and affordance are extended to include multiple timesteps prediction in the MDPs. Here we propose to extend the theory by generalise the definition of intent to capture \textit{an arbitrary kind of consequence} of an action taken at a state, generalising beyond state prediction. Such intents could be called \textit{general intent}. For example, the intent of a grasping action may include the desired success rate, object dropping rate, the weight of water that can be held, etc. Subsequently, the affordance is defined to include a subset of state and action pairs in which the intent is satisfied. Such affordances may be called \textit{general affordances}.

More importantly, this direction is promising if a thorough mathematical definition is developed based on the RL framework. A set of new algorithms can be developed to infer actions according to the predictions of arbitrary action consequences, instead of simply system states. Similar to the dynamics-based affordances, general affordances can help in exploration, value function or policy learning, model learning and planning by constraining the action space, but with respect to arbitrary action consequences beyond state prediction. However, this is outside of the scope of this review, and much more future efforts are required to derive and experiment the theory.

\section{Acknowledgement}
Xintong Yang thanks the Chinese Scholarship Council (CSC) for providing the living stipend for his Ph.D. programme (No. 201908440400). This work was partially supported by the Engineering and Physical Sciences Research Council (grant No. EP/X018962/1).

\printbibliography

\end{document}